\pdfoutput=1

\documentclass[11pt]{article}

\usepackage{emnlp2021}

\usepackage{times}
\usepackage{latexsym}
\usepackage{amsmath}
\usepackage{verbatim}
\usepackage{booktabs}
\usepackage{bm}
\usepackage{amssymb}
\usepackage{pifont}
\usepackage{stfloats}
\usepackage{multicol}
\usepackage{float}
\usepackage{graphicx}
\usepackage{subfigure}
\usepackage{multirow}
\usepackage{array}
\newcommand{\PreserveBackslash}[1]{\let\temp=\\#1\let\\=\temp}
\newcolumntype{C}[1]{>{\PreserveBackslash\centering}p{#1}}
\newcolumntype{R}[1]{>{\PreserveBackslash\raggedleft}p{#1}}
\newcolumntype{L}[1]{>{\PreserveBackslash\raggedright}p{#1}}
\usepackage{diagbox}
\usepackage[noend]{algpseudocode}
\usepackage{algorithmicx,algorithm}
\usepackage{mathtools}

\usepackage[T1]{fontenc}

\usepackage[utf8]{inputenc}

\usepackage{microtype}

\title{Modeling Concentrated Cross-Attention for Neural Machine Translation with Gaussian Mixture Model}

\author{Shaolei Zhang \textsuperscript{\rm 1,2},
    Yang Feng \textsuperscript{\rm 1,2}\thanks{ $\;\;$Corresponding author: Yang Feng.} \\
        \textsuperscript{\rm 1}{Key Laboratory of Intelligent Information Processing} \\ Institute of Computing Technology, Chinese Academy of Sciences (ICT/CAS) \\
    { \textsuperscript{\rm 2} {University of Chinese Academy of Sciences, Beijing, China}} \\
   \texttt{\{zhangshaolei20z, fengyang\}@ict.ac.cn}  }

\begin{document}
\maketitle
\begin{abstract}

Cross-attention is an important component of neural machine translation (NMT), which is always realized by dot-product attention in previous methods. However, dot-product attention only considers the pair-wise correlation between words, resulting in dispersion when dealing with long sentences and neglect of source neighboring relationships. Inspired by linguistics, the above issues are caused by ignoring a type of cross-attention, called concentrated attention, which focuses on several central words and then spreads around them. In this work, we apply Gaussian Mixture Model (GMM) to model the concentrated attention in cross-attention. Experiments and analyses we conducted on three datasets show that the proposed method outperforms the baseline and has significant improvement on alignment quality, N-gram accuracy, and long sentence translation.

\end{abstract}

\section{Introduction}
Recently, Neural Machine Translation (NMT) has been greatly improved with Transformer \cite{NIPS2017_7181}, which mainly relies on the attention mechanism. The attention mechanism in Transformer consists of self-attention and cross-attention, where cross-attention is proved more important to translation quality than self-attention \cite{2020arXiv201211747H,hard-code}. Even if the self-attention is modified to a fixed template, the translation quality would not significantly reduce \cite{hard-code}, while cross-attention plays an irreplaceable role in NMT. Cross-attention in Transformer is realized by dot-product attention, which calculates the attention distribution base on the pair-wise similarity.



However, modeling cross-attention with the dot-product attention still has some weaknesses due to its calculation way. First, when dealing with long sentences, the attention distribution with dot-product attention tends to be dispersed \cite{yang-etal-2018-modeling}, which proved unfavorable for translation \cite{8550728,tang-etal-2019-understanding,2012.11747}. Second, dot-product attention is difficult to explicitly consider the source neighboring relationship \cite{Distance-based-2017,Self-Attentional-2018}, resulting in ignoring the words with lower similarity but nearby the important word which determine phrase structure or morphology.


\begin{figure}[t]
\centering
\includegraphics[width=3in]{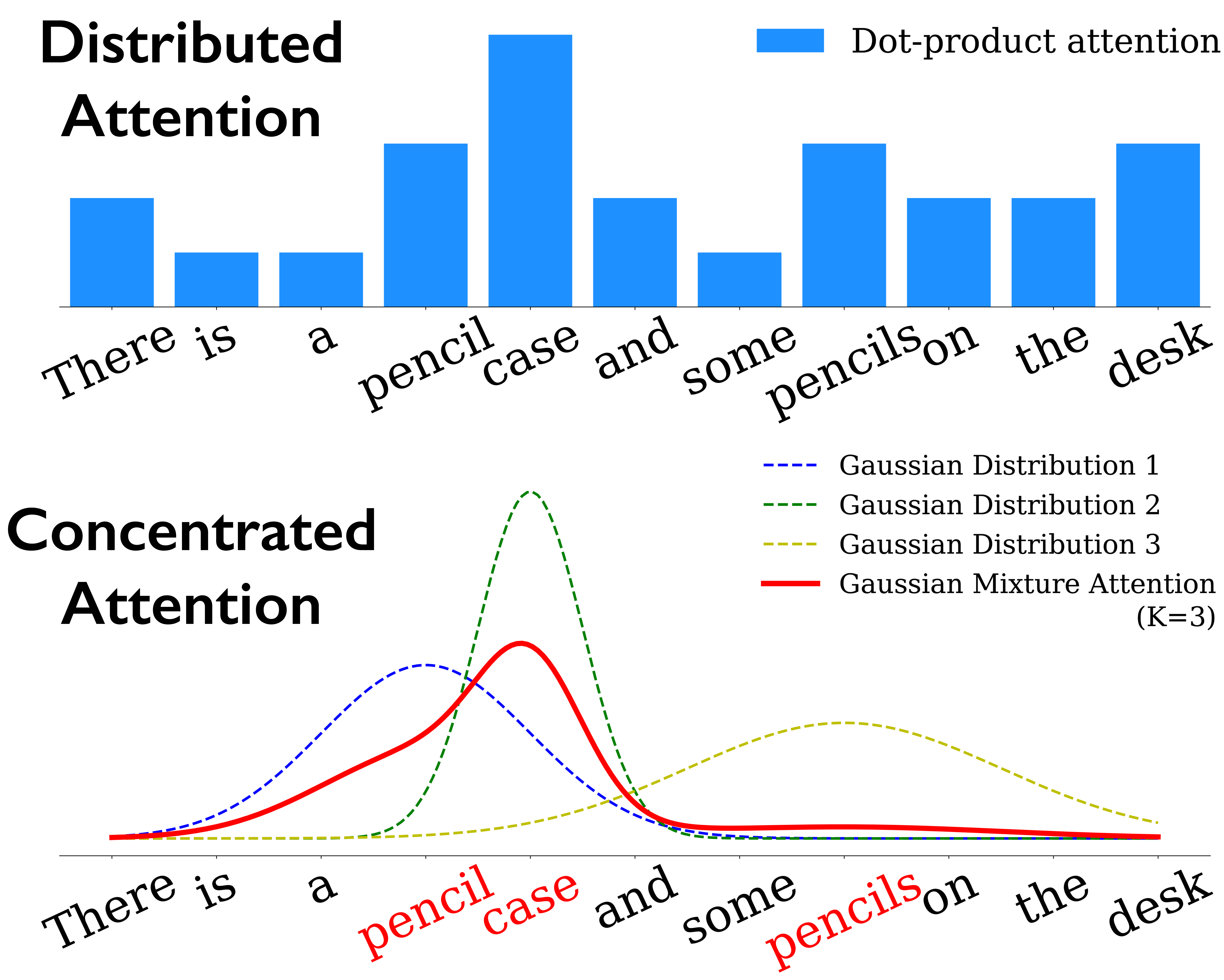}
\caption{An attention example of En$\rightarrow$De translation when generating target words ``Mäppchen'' (English meaning: pencil case), showing the difference and complementarity between distributed attention and concentrated attention. 
}
\label{show case}
\end{figure}



%

Research in linguistics and cognitive science suggests that human attention to language can be divided into two categories: \emph{distributed attention} and \emph{concentrated attention} \cite{Beck1973,ITO19981191,doi:10.1080/13506285.2018.1485808}. Specifically, distributed attention is scattered on all source words, and the degree of attention is determined by correlation. On the contrary, concentrated attention only focuses on a few central words and then spreads on the words around them. Accordingly, we consider that cross-attention can be divided into these two types of attention as well, where distributed attention is well modeled by dot-product attention, but concentrated attention is ignored. Figure \ref{show case} shows an attention example of En$\rightarrow$De translation when generating target words ``Mäppchen'' (English meaning: pencil case). Only with distributed attention, some irrelevant words (such as `desk') robbed some attention weight, resulting in attention dispersion. Besides, the correlation to the function words (both `a' and `some') are low and similar, but they are important to determine the singular/plural forms of the target word. In concentrated attention, attention distribution is more concentrated and can capture neighboring relationships, which make up for the lack of distributed attention.

In this paper, to explicitly model the concentrated cross-attention, we apply the Gaussian mixture model (GMM) \cite{DBLP:conf/nips/Rasmussen99} to construct \emph{Gaussian mixture attention}. Specifically, Gaussian mixture attention first focuses on some central words, and then pays attention to the words around the central words, where the attention decreases as the word away from the central word. Since cross-lingual alignments are often one-to-many, Gaussian mixture attention is more flexible to model multiple central words, which is not possible with a single Gaussian distribution. Then, Gaussian mixture attention and dot-product attention are fused to jointly determine the total attention.

Experiments we conducted on three datasets show that our method outperforms the baseline on translation quality. Further analyses show that our method enhances cross-attention, thereby improving the performance of alignment quality, N-gram accuracy and long sentence translation.

Our contributions can be summarized as follows:
\begin{itemize}
\setlength{\itemsep}{0pt}
\setlength{\parsep}{0pt}
\setlength{\parskip}{0pt}
    \item We introduce concentrated attention to cross-attention, which successfully compensates for the weakness of dot-product attention.
    \item To our best knowledge, we are the first to apply GMM to model attention distribution in text sequence, which provides a method for modeling multi-center attention distribution.
\end{itemize}

\section{Background}
Our method is applied on cross-attention in Transformer \cite{NIPS2017_7181}, so we first briefly introduce the architecture of Transformer with a focus on its dot-product attention. Then, we give the concept of the Gaussian Mixture model.

\subsection{Transformer}

\textbf{Encoder-Decoder} Transformer consists of encoder and decoder, each of which contains $N$ repeated independent structures. Each encoder layer contains two sub-layers: self-attention and fully connected feed-forward network (FFN), while each decoder layer includes three sub-layers: self-attention, cross-attention, and FFN. We denote the input sentence as $\mathbf{x}=\left ( x_{1} , \cdots ,x_{J}\right )$, where $J$ is the length of source sentence, $x_{j}\in \mathbb{R}^{d_{model}}$ is the sum of the token embedding and the position encoding of the source token, and $d_{model}$ represents the representation dimension. The encoder maps $\mathbf{x}$ to a sequence of hidden states $\mathbf{z}=\left ( z_{1} , \cdots ,z_{J}\right )$. Given $\mathbf{z}$ and previous target tokens, the decoder predicts the next output token $y_{i}$, and the entire output sequence is $\mathbf{y}=\left ( y_{1} , \cdots ,y_{I}\right )$, where $I$ is the length of the target sentence.

\textbf{Dot-product attention} Both self-attention and cross-attention in Transformer apply multi-head attention, which contains multiple heads and each head accomplishes scaled dot-product attention to process a set of queries(q), keys(k), and values(v). Following, we focus on the specific representation of the dot-product attention in cross-attention.

In cross-attention, the queries is the hidden states of decoder $\mathbf{s}=\left \{ s_{1},\cdots ,s_{I} \right \}$, while the keys and values both come from the hidden states of the encoder $\mathbf{z}=\left \{ z_{1},\cdots ,z_{J} \right \}$. 
Dot-product attention first calculates the pairwise correlation score $e_{ij}$ between the $i^{th}$ target token and the $j^{th}$ source token, and normalizes to obtain the dot-product attention weight $\alpha _{ij}$ of the $i^{th}$ target token for each source token:
\begin{align}
e_{ij}&=\frac{Q\left ( s_{i} \right )K\left ( z_{j} \right )^{T}}{\sqrt{d_{k}}}\label{eq3} \\
\alpha _{ij}&=\frac{ \exp e_{ij}}{\sum_{l=1}^{J}\exp e_{il}}\label{eq4}
\end{align}
where $Q\left ( \cdot  \right )$ and $K\left ( \cdot  \right )$ are the projection functions from the input space to the query space and the key space, respectively, and $d_{k}$ represents the dimensions of the queries and keys. 
Then for each $i$, the context vector $c_{i}$ is calculated as:
\begin{gather}
c_{i}=\sum_{j=1}^{n} \alpha _{ij}V\left ( z_{j} \right ) \label{eq2}
\end{gather}
where $V\left ( \cdot  \right )$ is a projection function from the input space to the value space.

\subsection{Gaussian Mixture Model}

\textbf{Single Gaussian distribution} A Gaussian distribution with the mean $\mu$ and variance $\sigma$, which is calculated as:
\begin{equation}
    x\sim \mathcal{N}\left ( \mu ,\sigma  \right )\equiv \frac{1}{\sqrt{2\pi }\sigma}\mathrm{exp}\left ( -\frac{\left ( x-\mu \right )^{2}}{2\sigma ^{2}} \right )
\end{equation}

\textbf{Gaussian mixture model (GMM)} \cite{DBLP:conf/nips/Rasmussen99} Composed of $K$ single Gaussian distribution, which is calculated as:
\begin{equation}
    x\sim \sum_{k=1}^{K}a_{k}\mathcal{N}\left ( \mu_{k} ,\sigma_{k}  \right )
\end{equation}
where $a_{k}$, $\mu_{k}$ and $\sigma_{k}$ are weight, mean and variance of the $k^{th}$ Gaussian distribution, respectively. During training, for unlabeled data, GMM can be trained using the EM algorithm, and for labeled data, GMM can be trained using methods such as maximum likelihood or gradient descent.

\section{The Proposed Method}
\label{sec:Method}

\begin{figure}[t]
\includegraphics[width=3in]{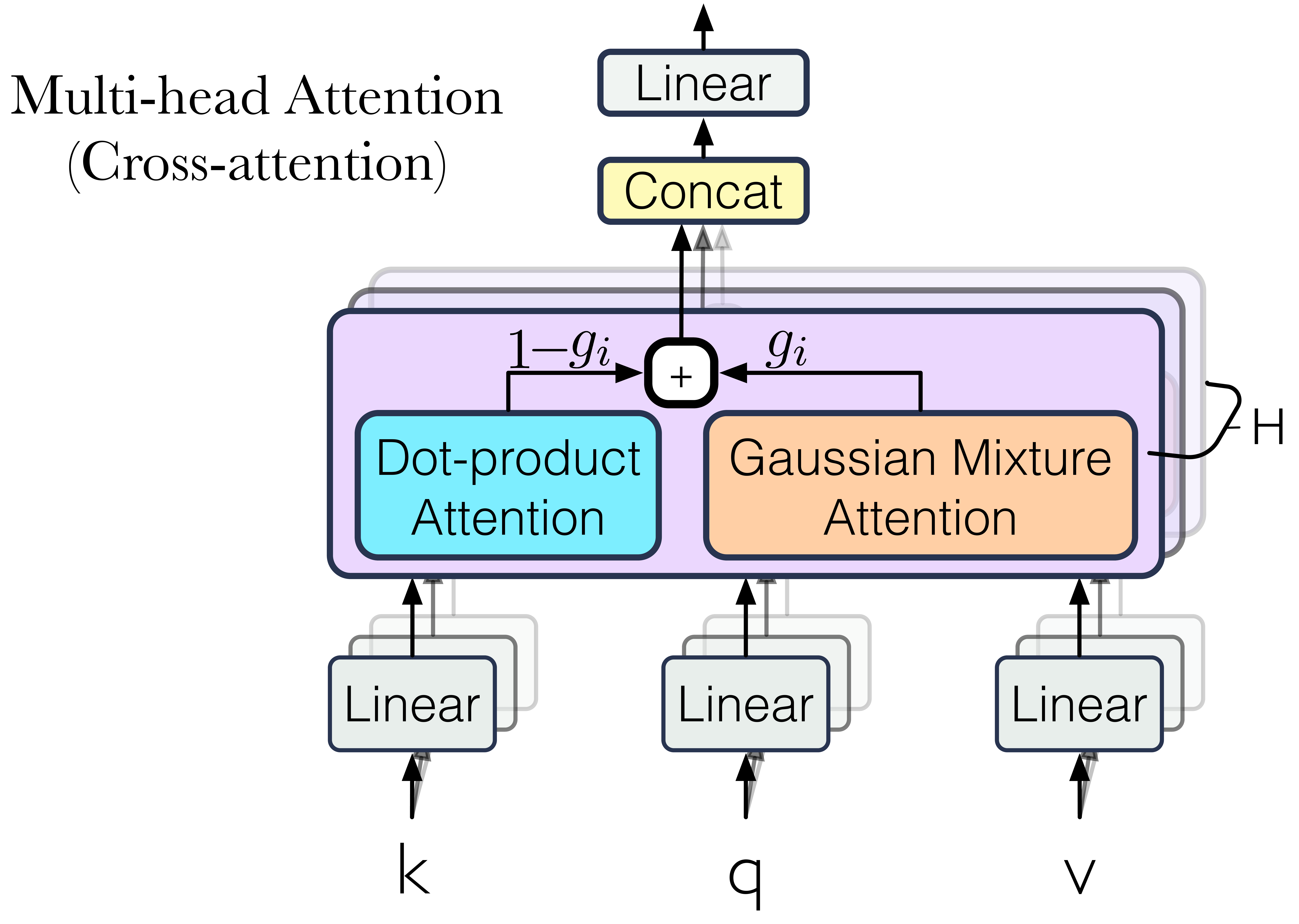}
\caption{The architecture of the proposed method.}
\label{multi}
\end{figure}

To improve cross-attention, in addition to dot-product attention, we introduce the Gaussian mixture attention into each head of cross-attention. As shown in Figure \ref{multi}, we first calculate the dot-product attention and Gaussian mixture attention, and then fuses them through a gating mechanism to determine the total attention distribution. The proposed Gaussian mixture attention is constructed by mean, variance, and weight, all of which are predicted based on the target word, as shown in Figure \ref{model}. The specific details will be introduced following.

\subsection{Gaussian Mixture Attention}
\label{sec:GMM}

Gaussian mixture attention consists of $K$ independent Gaussian distributions, where each Gaussian distribution has different weights. The general form of Gaussian mixture attention $\beta _{ij}$ between the $i^{th}$ target token and the $j^{th}$ source token is defined as:
\begin{equation}
    \beta _{ij}= \sum_{k=1}^{K}\omega _{ik}\cdot \frac{1}{Z_{ik}}\mathrm{exp}\left ( -\frac{\left ( j-\mu _{ik} \right )^{2}}{2\sigma _{ik}^{2}} \right ) \label{eq8}
\end{equation}
where $\mu _{ik}$, $\sigma _{ik}$ and $\omega _{ik}$ are the mean, variance, and the weight of the $k^{th}$ Gaussian distribution of the $i^{th}$ target word, respectively, $Z_{ik}$ represents the normalization factor of $k^{th}$ Gaussian distribution of the $i^{th}$ target word, and $K$ is a hyperparameter we set. In the physical sense, the mean $\mu _{ik}$ represents the position of the central word, the variance $\sigma _{ik}$ represents the attenuation degree of attention with the offset from the central word, and the weight $\omega _{ik}$ represents the importance of the central word.

\begin{figure}[t]
\centering
\includegraphics[width=2.02in]{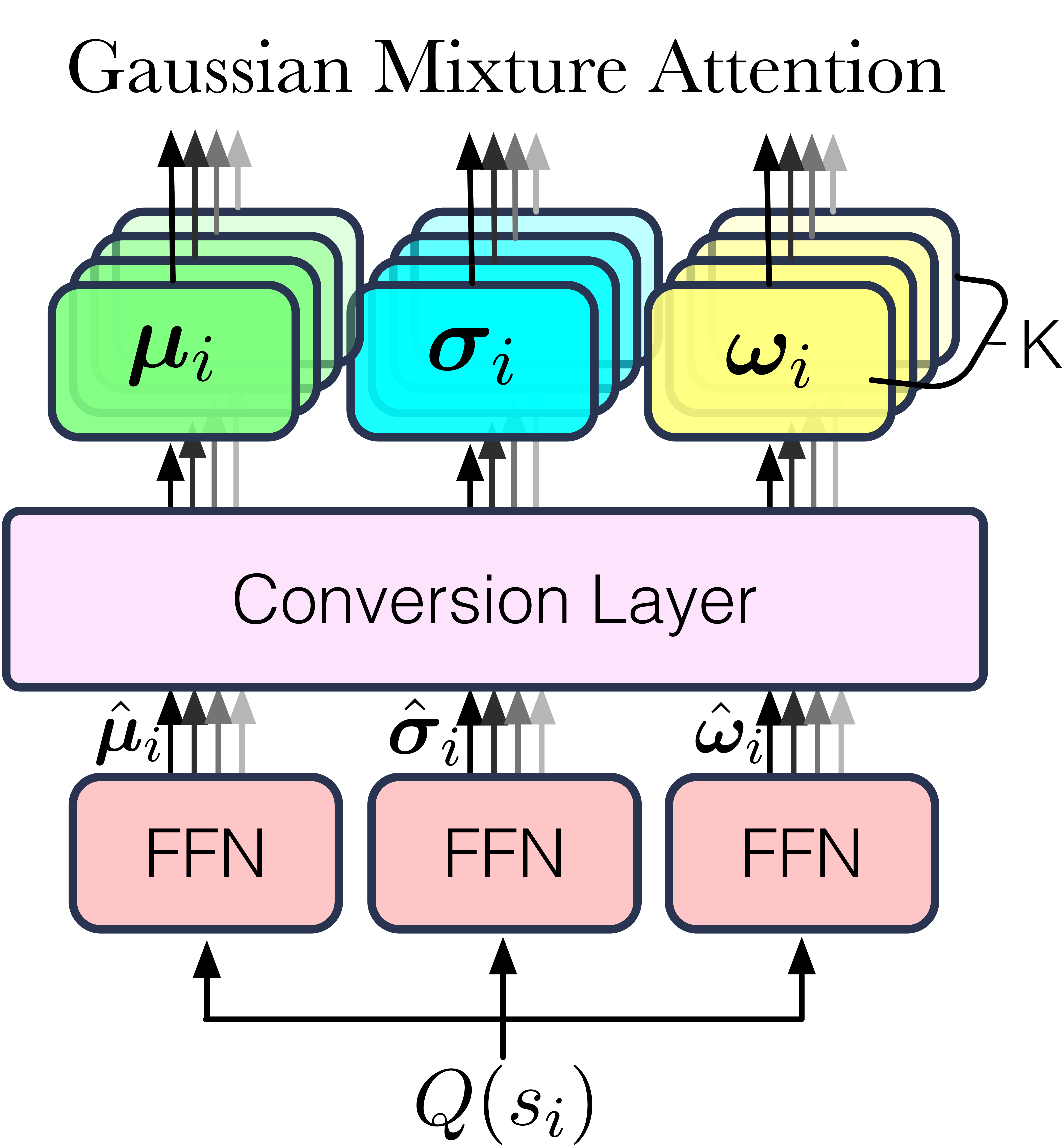}
\caption{Calculation of Gaussian Mixture Attention.}
\label{model}
\end{figure}

As shown in Figure \ref{model}, the parameters for the Gaussian mixture attention of the $i^{th}$ target word are $(  \bm{\omega }_{i} , \bm{\mu }_{i} , \bm{\sigma }_{i},\bm{Z}_{i})$, where $\bm{\omega }_{i}\in \mathbb{R}^{K \times 1}$ is the vector representation of $\left [ \omega_{i1},\cdots , \omega_{iK} \right ]$, and others are in the same rule.
$(  \bm{\omega }_{i} , \bm{\mu }_{i} , \bm{\sigma }_{i},\bm{Z}_{i})$ are converted from predicted intermediate parameters $(  \hat{\bm{\omega }}_{i} , \hat{\bm{\mu }}_{i} , \hat{\bm{\sigma }}_{i})$. According to \citet{yang-etal-2018-modeling} and \citet{9054106}, it is more rubost to use target hidden state to predict variables of Gaussian distribution. Thus, the intermediate parameters are predicted through the Feedforward Network (FFN):
\begin{gather}
   \hat{\bm{\omega }}_{i}  =\bm{V}_{\omega }^{\top }\, \mathrm{tanh}\left ( \bm{W}_{\omega }^{\top }Q\!\left ( s_{i} \right )+b_{\omega1 } \right )+b_{\omega2 } \\
   \hat{\bm{\mu }}_{i}  =\bm{V}_{\mu }^{\top }\, \mathrm{tanh}\left ( \bm{W}_{\mu }^{\top }Q\!\left ( s_{i} \right )+b_{\mu1 } \right )+b_{\mu2 } \\
   \hat{\bm{\sigma }}_{i}  =\bm{V}_{\sigma }^{\top }\, \mathrm{tanh}\left ( \bm{W}_{\sigma }^{\top }Q\!\left ( s_{i} \right )+b_{\sigma1 } \right )+b_{\sigma2 } 
\end{gather}
where $\left \{\!\bm{W}_{\omega },\! \bm{W}_{\mu },\! \bm{W}_{\sigma }\!\right \} $$\in$$ \mathbb{R}^{d_{q}\!\times\! d_{q}} $ and $\left \{\!\bm{V}_{\omega },\! \bm{V}_{\mu },\! \bm{V}_{\sigma }\!\right \} $
$\in$$ \mathbb{R}^{d_{q}\!\times\! K} $ are learnable parameters, $\left \{b_{\omega1 },\! b_{\mu1 },\! b_{\sigma1 }\right \}$
$\in$$ \mathbb{R}^{d_{q}\!\times\! 1} $ and $\left \{b_{\omega2 },\! b_{\mu2 },\! b_{\sigma2 }\right \}$$\in$$ \mathbb{R}^{K\!\times\! 1} $ are learnable bias. Note that $Q\left ( \cdot  \right )$ shares parameters with the projection function of dot-product attention in Eq.(\ref{eq4}), and parameters of FFN are shared in each head.

Given intermediate parameters $(  \hat{\bm{\omega }}_{i} , \hat{\bm{\mu }}_{i} , \hat{\bm{\sigma }}_{i})$, our method predicts $(  \bm{\omega }_{i} , \bm{\mu }_{i} , \bm{\sigma }_{i},\bm{Z}_{i})$ through a conversion layer. For the weight of every single Gaussian distribution, we normalize them with:
\begin{equation}
    \bm{\omega }_{i}=\mathrm{Softmax}\left ( \hat{\bm{\omega }}_{i} \right )
\end{equation}
For the mean, considering the word order differences between language, we directly predict its absolute position:
\begin{equation}
    \bm{\mu }_{i}=J\cdot \mathrm{Sigmoid}\left ( \hat{\bm{\mu }}_{i}  \right )
\end{equation}
where $J$ is the length of the source sentence. 

Note that, since the source position is discrete and will be truncated at the boundary, the attention weight sum is less than 1 without normalization, rather than fully normalized attention weight. Previous work \cite{luong-etal-2015-effective,yang-etal-2018-modeling,hard-code} on applying Gaussian distribution hardly normalized it since the normalization of Gaussian attention weights results in unstable training. However, unnormalized attention weight leads to occasional spikes or dropouts in the attention and alignment \cite{9054106}. Therefore, to normalize Gaussian mixture attention meanwhile maintaining training stability, we propose an approximate normalization.

Approximate normalization adjusts the variance according to the mean position to ensure that most of the weights are within the source sentence, which not only avoids the spikes caused by little weights sum but also ensures stable training. For approximate normalization, we calculate the value of $\bm{\sigma }_{i}$ with $\bm{\mu }_{i}$:
\begin{gather}
        \bm{\sigma }_{i}\!=\!\mathrm{min}\left \{\frac{J}{6}\cdot \mathrm{Sigmoid}\left ( \hat{\bm{\sigma }}_{i}  \right ),\!\frac{\bm{\mu }_{i}}{3},\!\frac{J\!-\!\bm{\mu }_{i}}{3} \right \} \label{eq13}
\end{gather}
Approximate normalization ensures that interval $\left [\bm{\mu }_{i}-3\bm{\sigma }_{i},\bm{\mu }_{i}+3\bm{\sigma }_{i}  \right ]$ is within the source sentence \cite{3sigma}. Besides, $\bm{Z}_{i}$ is set to $\sqrt{2\pi \bm{\sigma }_{i}^{2}}$ to normalize each Gaussian distribution in GMM. In general, although Gaussian mixture attention is not strictly normalized, approximate normalization guarantees a coverage of more than 90\% of the attention weight. In the experiments (Sec.\ref{sec:system}), we additionally report the results of a strict normalization Gaussian mixture attention as a variant of our method for comparison.

\subsection{Fusion of Attention}

Dot-product attention ($\alpha _{ij}$ in Eq.(\ref{eq4})) capture the distributed attention brought by the pair-wise similarity, while Gaussian mixture attention ($\beta _{ij}$ in Eq.(\ref{eq8})) model the location-related concentrated attention. To balance two types of attention, we calculate the total attention weight $\gamma _{ij}$ by fusing them through a gating mechanism:
\begin{equation}
    \gamma _{ij}=\left ( 1-g_{i} \right )\times  \alpha _{ij} + g_{i}\times \beta _{ij} \label{eq14}
\end{equation}
where $g_{i}$ is a gating factor, predicted through FFN:
\begin{equation}
    g_{i}  \!=\!\mathrm{Sigmoid}(\bm{V}_{g }^{\top }\mathrm{tanh}\left ( \!\bm{W}_{g}^{\top }Q\!\left ( s_{i} \right )\!+\!b_{g1 } \right )\!+\!b_{g2 })
\end{equation}
where $\bm{W}_{g }\in \mathbb{R}^{d_{q}\times d_{q}} $ and $\bm{V}_{g }\in \mathbb{R}^{d_{q}\times 1} $ are learnable parameters, $b_{g1 }\in \mathbb{R}^{d_{q}\times 1} $ and $b_{g2 }\in \mathbb{R}$ are learnable bias. Finally, the context vector $c_{i}$ in Eq.(\ref{eq2}) is calculated as:
\begin{gather}
c_{i}=\sum_{j=1}^{n} \gamma _{ij}V\left ( z_{j} \right )
\end{gather}

\section{Related Work}

Attention mechanism is the most significant component of the Transformer \cite{NIPS2017_7181} for Neural Machine Translation. Recently, some methods model the location in Transformer, most of which focus on self-attention.

Some methods improve the word representation. Transformer itself \cite{NIPS2017_7181} introduced a position encoding to embed position information in word representation. \citet{shaw-etal-2018-self} introduced relative position encoding in self-attention. \citet{wang-etal-2019-self-attention} enhanced self-attention with structural positions from the syntax dependencies. \citet{ding-etal-2020-self} utilized reordering information to learn position representation in self-attention.

Other methods directly consider location information in the attention mechanism, which is closely related to this work. \citet{luong-etal-2015-effective} proposed local attention, which only focuses on a small subset of the source positions. \citet{yang-etal-2018-modeling} multiplied a learnable Gaussian bias to self-attention to model the local information. \citet{1811.00253} accommodates some masks for self-attention to extract global/local information. \citet{xu-etal-2019-leveraging} propose a hybrid attention mechanism to dynamically leverage both local and global information. \citet{hard-code} apply a hard-code Gaussian to replace dot-product attention in Transformer.

There are three differences between the proposed method and previous methods. 1) Most previous methods only focus on self-attention, while we consider the cross-attention, which is proved to be more critical to translation quality \cite{voita-etal-2019-analyzing,tang-etal-2019-understanding,hard-code,ding-etal-2020-context}. 2) Previous methods usually multiply dot-product attention with a position-related bias or mask to model position. Our method additionally introduces a concentrated attention to compensate for dot-product attention, rather than simple bias. 3) Gaussian distribution is widely used in previous position modeling, while we use the more flexible GMM for complex cross-attention.

\section{Experiments}

\begin{table*}[]
\centering
\begin{tabular}{L{3.4cm}|C{1cm}C{2cm}C{2.5cm}C{2.5cm}|c}\toprule[1pt]
\textbf{Models}   & $\bm{Z}_{i}$ & $\bm{\omega }_{i}$ & $\bm{\mu }_{i}$ & $\bm{\sigma }_{i}$ & \textbf{Normalized} \\  \midrule
 \textbf{Synthesis Network} & $\bm{1}$  & $\mathrm{exp}\left ( \hat{\bm{\omega }}_{i} \right )$  &  $\bm{\mu }_{i-1}+\mathrm{exp}\left ( \hat{\bm{\mu }}_{i} \right )$     &  $\sqrt{\mathrm{exp}\left ( -\hat{\bm{\sigma }}_{i} \right )/2} $       &   None         \\\midrule[0.1pt]
\textbf{Our Method} & $\sqrt{2\pi \bm{\sigma }_{i}^{2}}$  & $\mathrm{Softmax}\left ( \hat{\bm{\omega }}_{i} \right )$  &  $J\cdot \mathrm{Sigmoid}\left ( \hat{\bm{\mu }}_{i}  \right )$     &    Eq.(\ref{eq13})       &    Approximate       \\
 \textbf{Our Method+Norm.} & $\sqrt{2\pi \bm{\sigma }_{i}^{2}}$  & $\mathrm{Softmax}\left ( \hat{\bm{\omega }}_{i} \right )$  & $J\cdot \mathrm{Sigmoid}\left ( \hat{\bm{\mu }}_{i}  \right )$      &  $J\cdot \mathrm{Sigmoid}\left ( \hat{\bm{\sigma }}_{i}  \right )$        &   Strict        \\
\bottomrule[1pt]
\end{tabular}
\caption{Conversion method of synthesis network, our method, and the strict-normalized variant of our method.}
\label{version}
\end{table*}
We conducted experiments on three datasets and compare with the baseline and previous methods to evaluate the performance of the proposed method.

\subsection{Datasets}
Experiments were conducted on the following three datasets of different sizes.

\textbf{Nist Zh$\rightarrow $En} 1.25M sentence pairs from LDC corpora\footnote{The corpora include LDC2002E18, LDC2003E07, LDC2003E14, Hansards portion of LDC2004T07, LDC2004T08 and LDC2005T06.}. We use MT02 as the validation set and MT03, MT04, MT05, MT06, MT08 as the test sets, each with 4 English references. Results are averaged on all test sets. We tokenize and lowercase English sentences with the Moses\footnote{\url{https://www.statmt.org/moses/}}, and segmented the Chinese sentences with Stanford Segmentor\footnote{\url{https://nlp.stanford.edu/}}. We apply BPE \cite{sennrich-etal-2016-neural} with 30K merge operations on all texts.


\textbf{WMT14 En$\rightarrow $De}  4.5M sentence pairs from WMT14 \footnote{\url{www.statmt.org/wmt14/}} English-German task. We use news-test2013 (3000 sentence pairs) as the validation set and news-test2014 (3003 sentence pairs) as the test set. We apply BPE with 32K merge operations, and the vocabulary is shared across languages.

\textbf{WMT17 Zh$\rightarrow $En} 20M sentence pairs from WMT17 \footnote{\url{www.statmt.org/wmt17/}} Chinese-English task, follow the processing of \citet{Hassan2018AchievingHP}. We use devtest2017 (2002 sentence pairs) as the validation set and news-test2017 (2001 sentence pairs) as the test set. We apply BPE with 32K merge operations on all texts.


\subsection{System}
\label{sec:system}

We conducted experiments on the following systems.

{\bf {Transformer}} Baseline of our method. The architecture of {Transformer}-\textit{Base}/\textit{Big} was implemented strictly referring to \citet{NIPS2017_7181}. 

{\bf {Hard-Code Gaussian}} Use a hard-code Gaussian distribution to replace dot-product attention in cross-attention \cite{hard-code}. The hard-code Gaussian distribution is an artificially set Gaussian distribution with fixed mean and variance.

{\bf{Localness Gaussian}} Our implementation of the modeling localness proposed by \citet{yang-etal-2018-modeling}. A learnable Gaussian bias is multiplied to model the local information of attention, especially in self-attention, and we apply it in cross-attention.

{\bf {Synthesis Network}} Following \cite{journals/corr/Graves13}, We modify the proposed Gaussian mixture attention with the synthesis network, which consists of $K$ unnormalized Gaussian distributions, as shown in Table \ref{version}.

{\bf {Our Method}} The proposed method. A Gaussian mixture attention is applied to cross-attention in Transformer. Refer to Sec.\ref{sec:Method} for specific details.

{\bf {Our Method+Norm.}} Considering our method is approximate-normalized, we additionally propose strict-normalized Gaussian mixture attention, as a variant of our method. Since Gaussian mixture attention weights $\beta _{ij}$ are all positive numbers, we directly use $\beta _{ij}/\sum_{l=1}^{n}\beta _{ij}$ to normalize it, referring to `Our Method+Norm.' in Table \ref{version}.

The implementation is all adapted from Fairseq Library \cite{ott-etal-2019-fairseq} with the same settings from \citet{NIPS2017_7181}. SacreBLEU \cite{post-2018-call} is applied to evaluate translation quality.
\begin{table*}[]
\centering
\begin{tabular}{L{4.2cm}|C{1.4cm}R{1.4cm}|C{1.4cm}R{1.4cm}|C{1.4cm}R{1.4cm}} \toprule[1pt]
{\multirow{2}{*}{\textbf{Models}}}             & \multicolumn{2}{c|}{\textbf{\begin{tabular}[c]{@{}c@{}}Nist\\ Zh$\rightarrow $En\end{tabular}}}  & \multicolumn{2}{c|}{\textbf{\begin{tabular}[c]{@{}c@{}}WMT14\\ En$\rightarrow $De\end{tabular}}} & \multicolumn{2}{c}{\textbf{\begin{tabular}[c]{@{}c@{}}WMT17\\ Zh$\rightarrow $En\end{tabular}}} \\ \cline{2-7}
            & BLEU & \#Para.   & BLEU    & \#Para.  & BLEU    & \#Para.  \\ \midrule\midrule
{\textbf{{Transformer}-\textit{Base}}}   & 44.02    &   79.7M                                          & 27.34   &  63.1M                      & 24.03   &     83.6M                   \\
{\textbf{Hard-Code Gaussian}}  &36.10     &  79.6M                                            & 25.03    &  63.0M                      &20.95    & 83.5M                       \\
{\textbf{Localness Gaussian}}  & 44.06    & 80.2M                                            & 27.41   & 63.6M                       & 24.01   & 84.1M                       \\ 
{\textbf{Synthesis Network}}  & 43.34    & 79.8M                                  & 26.27   & 63.2M                 & 23.69   & 83.7M                   \\ \midrule[0.1pt]
\textbf{Our Method+Norm.}          & 44.69$\uparrow$   & 79.8M                                     & 27.50$\uparrow$   & 63.2M                   & 24.44$\uparrow$   & 83.7M                   \\
\textbf{Our Method}          & \textbf{45.39}$\uparrow$    & 79.8M                                    & \textbf{28.09}$\uparrow$   & 63.2M                   & \textbf{24.44}$\uparrow$   & 83.7M        \\   \midrule\midrule
{\textbf{{Transformer}-\textit{Big}}} &44.20     &  247.5M                                         & 28.43   & 214.3M                       &24.46    &  255.2M                      \\ \midrule[0.1pt]
\textbf{Our Method (\textit{Big})}          &\textbf{45.45}$\uparrow$     & 247.6M                                & \textbf{29.02}$\uparrow$   & 214.4M                 &  \textbf{24.82}$\uparrow$   &  255.3M          \\ \bottomrule[1pt]
\end{tabular}
\caption{BLEU score of our method and the existing NMT models on test sets. ``\#Para.'': the learnable parameter scale of the model (M=million). ``$\uparrow$'': the improvement is signiﬁcant by contrast to baseline ($\rho < 0.01$).}
\label{main}
\end{table*}
\subsection{Effect of Hyperparameter $K$}

\begin{figure}[t]
\includegraphics[width=3in]{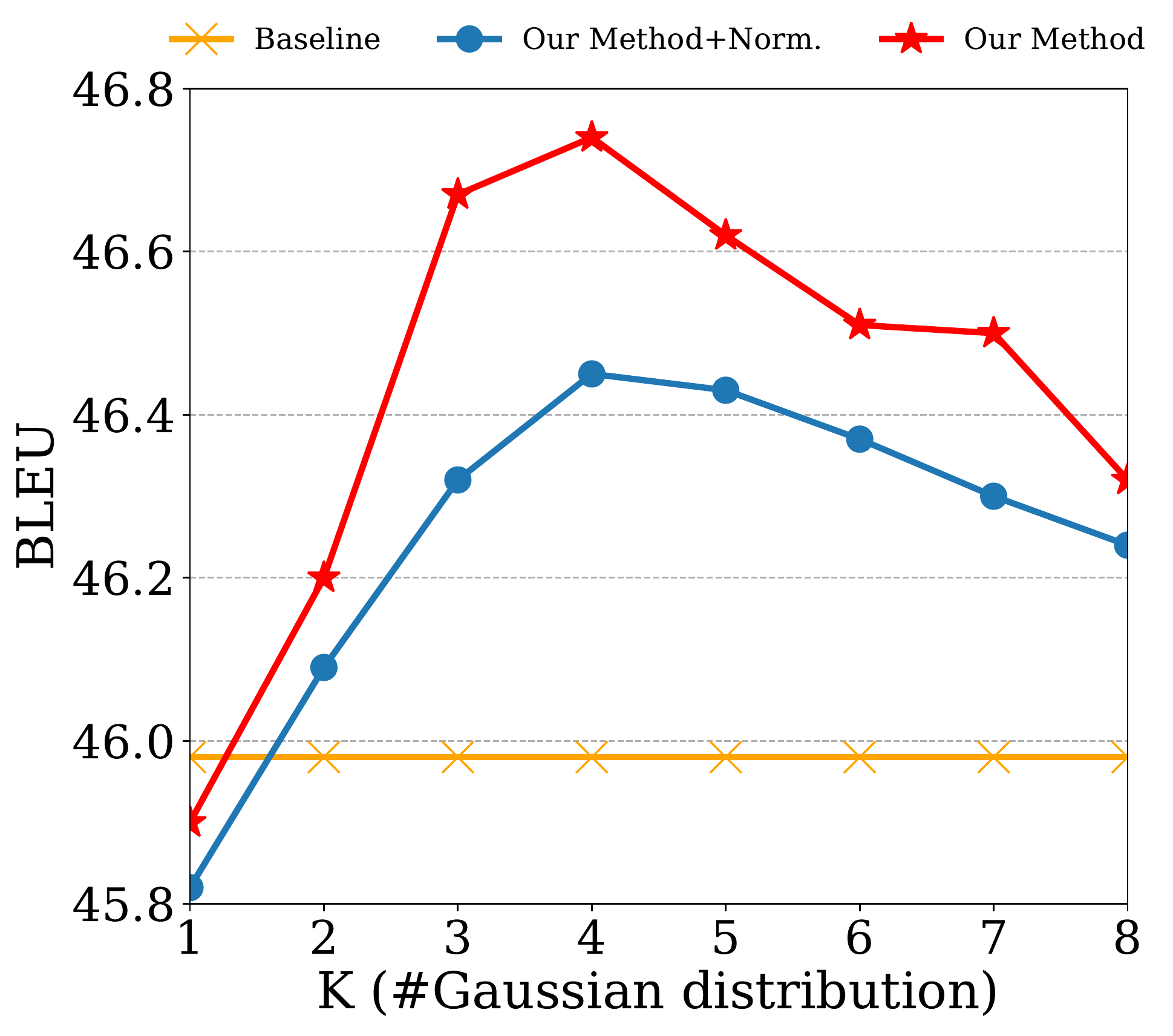}
\caption{BLEU scores with different $K$.}
\label{k-valid}
\end{figure}

Before the main experiment, as shown in Figure \ref{k-valid}, we evaluate performance with various hyperparameter $K$ on the Nist Zh$\rightarrow$En validation set, where $K$ represents the number of Gaussian distributions in the Gaussian mixture attention. When $K=1$, since the cross-attention is not one-to-one, it is difficult for a single Gaussian distribution to fit the cross-attention. With the increase of $K$, the translation quality improves and performs best when $K=4$. When $K$ is large, too many Gaussian distributions in GMM-based attention will complicate the model and maybe predict some inaccurate central words, resulting in a decrease in translation quality. Therefore, we set $K=4$ in the following experiments.

\subsection{Main Results}

Table \ref{main} shows the results of our method compared with the baseline and previous methods. Our method achieves the best results on all three datasets, improving about 1.37 on Nist Zh$\rightarrow$En, 0.75 on WMT14 En$\rightarrow $De, and 0.41 on WMT17 Zh$\rightarrow $En respectively, compared with Transformer-\textit{Base}. Besides, compared to Transformer-\textit{Big}, our method still brings significant improvements. Our method only increases 0.1\% more parameters than Transformer-\textit{Base} and achieve similar performance with Transformer-\textit{Big}. The performance improvement of our method is not simply through increasing the model parameters, but to improve the cross-attention with the proposed Gaussian mixture attention.

`Hard-Code Gaussian' \cite{hard-code} and `Localness Gaussian' \cite{yang-etal-2018-modeling} have been proved to be effective in self-attention but not obvious for cross-attention, since it's difficult to fit complex cross-attention with their single Gaussian bias. The Gaussian mixture attention is more flexible and can fit arbitrarily complex distributions, especially to model multi-center attention distribution, which is more suitable for modeling cross-attention.

Considering normalization, `Synthesis Network' is un-normalized, `Our Method+Norm.' is strict-normalized, and `our method' is approximate-normalized, where our proposed approximate normalization performs best. In practice, un-normalization tends to cause attention spikes, while strict normalization leads to unstable training.

\section{Analysis}

We conducted extensive analyses to understand the specific improvements of our method in attention entropy, alignment quality, phrase fluency, and long sentence translation. Unless otherwise specified, all the results are reported on WMT14 En$\rightarrow $De test set with {Transformer}-\textit{Base}.

\subsection{Flexibility of Gaussian Mixture Attention}

\begin{table}[]
\centering
\begin{tabular}{l|cc} \toprule[1pt]
                        & \textbf{BLEU} & \bm{$\Delta$} \\ \midrule
\textbf{Our Method}     & \textbf{28.09}         &                \\ \midrule[0.1pt]
\textbf{$\;\;\;\bm{-}\;$Share Mean}     & 27.58         & -0.51          \\
\textbf{$\;\;\;\bm{-}\;$Share Varience} & 27.67              &  -0.42              \\
\textbf{$\;\;\;\bm{-}\;$Share Weight}   & 27.90              &  -0.19             \\ \bottomrule[1pt]
\end{tabular}
\caption{Performance when each Gaussian distribution shares the mean, variance, and weight, respectively.}
\label{flex}
\end{table}

Compared with the single Gaussian distribution, GMM is more flexible on three aspects: mean, variance, and weight. To evaluate the improvements brought by these three aspects, we respectively share the mean, variance, and weight between each Gaussian distribution in Gaussian mixture attention, and report the results in Table \ref{flex}.

The performance decreases most obviously when each Gaussian distribution sharing the same central word (mean). The major superiority of GMM over Gaussian distribution is that GMM contains multiple centers, which is more in line with cross-attention. The variance allows each central word to have different attention coverage, and the weight controls the contribution of each Gaussian distribution. The flexibility of these three aspects makes Gaussian mixture attention more suitable for cross-attention.

\subsection{Effect of Gating Mechanism}

\begin{table}[]
\centering
\begin{tabular}{l|cc} \toprule[1pt]
            & \textbf{BLEU}  & \bm{$\Delta$} \\ \midrule

\textbf{Our Method}          & \textbf{28.09} &   \\
\textbf{$\;\;\;\;\;\bm{-}\;$Average Gating}   & 27.61 & -0.48  \\ \midrule[0.1pt]
\textbf{Dot-product Attention} & 27.34  &  -0.75     \\
\textbf{Gaussian Mixture Attention} & 26.32  & -1.77      \\ \bottomrule[1pt]
\end{tabular}
\caption{Ablation study of directly averaging the two types of attention or only using one type of attention.}
\label{wogate}
\end{table}

Our method applies a gating mechanism to fuse Gaussian mixture attention and dot-product attention. We conduct the ablation study of directly averaging the two types of attention or only using one of them in Table \ref{wogate}. Our method surpasses only using a single type of attention or directly averaging the two types of attention, which shows that the gating mechanism plays an important role and effectively fuses two types of attention.

\begin{figure}[t]
\centering
\includegraphics[width=3in]{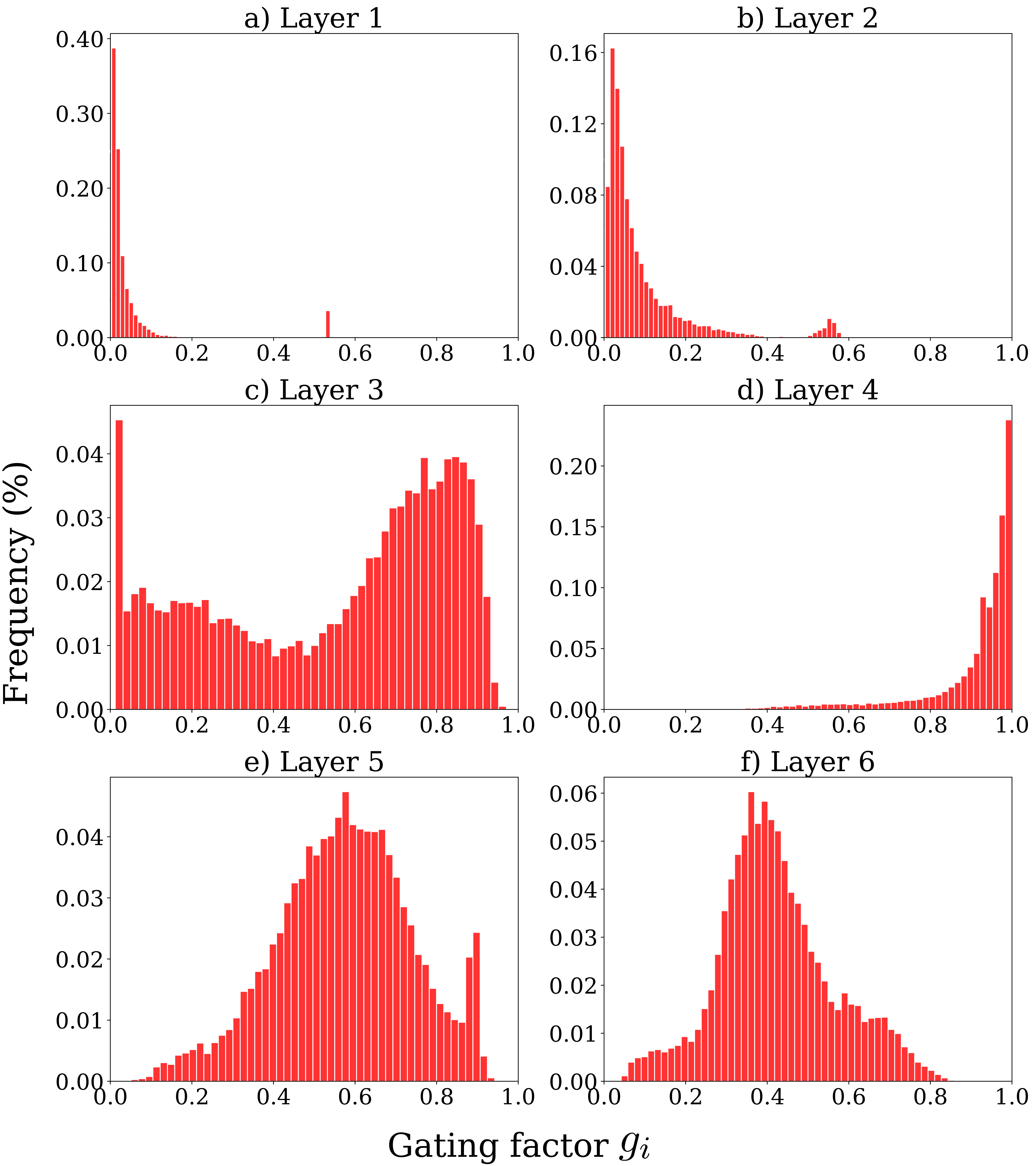}
\caption{The distribution of gating factor $g_{i}$ in the cross-attention of each layer. X-axis is $g_{i}$, which represents the weight of Gaussian mixture attention in the total attention, and Y-axis is the frequency with $g_{i}$.}
\label{gate}
\end{figure}

To analyze the relationship between the two types of attention in detail, we calculate the distribution of gating factor ($g_{i}$ in Eq.(\ref{eq14})) of each decoder layer, and show the result in Figure \ref{gate}. To our surprise, our method makes the cross-attention in each decoder layer present a different division of labor, which confirms the conclusions of previous work \cite{li-etal-2019-word}. Specifically, the bottom layers (L1, L2) in the decoder emphasize on dot-product attention and tend to capture global information; the middle layers (L3, L4) emphasize on Gaussian mixture attention, which captures local information around the central word; two types of attention in the top layer (L5, L6) are more balanced and jointly determine the final output. Previous work \cite{li-etal-2019-word} pointed out that the cross-attention of each layer is different, where the lower layer tends to capture the sentence information, while the upper layer tends to capture the alignment and specific word information since it is closer to the output. Our method also confirms this point, where Gaussian mixture attention effectively model concentrated attention so that it occupies a larger proportion in higher layers. With the gating mechanism, our method successfully fuses two types of attention and learns the division of labor between different layers.

Based on this, we tried to only apply our method to a part of decoder layers to verify the effect of our method on different layers, and the results are reported in Table \ref{layer}. When Gaussian mixture attention is only used in the top or middle 2 layers, the translation quality can be significantly improved without requiring many additional calculations compared with Transformer-\textit{Base}.

\begin{table}[]
\centering
\begin{tabular}{c|r|cc} \toprule[1pt]
\multicolumn{2}{c|}{}                                                                  & \textbf{BLEU}  & \bm{$\Delta$} \\\midrule
\multicolumn{2}{c|}{\textbf{Baseline}}                                                          & 27.34 &       \\ \midrule
\multirow{4}{*}{\begin{tabular}[c|]{@{}c@{}}\textbf{Our}|\\ \textbf{method}\end{tabular}} & All 6 layers      & 28.09 & +0.75 \\\cline{2-4}
                                                                      & Bottom 2 layers  & 27.43 & +0.09 \\
                                                                      & Middle 2 layers  & 27.74 & +0.40 \\
                                                                      & Top 2 layers & 27.89 & +0.55 \\\bottomrule[1pt]
\end{tabular}
\caption{Result of using Gaussian mixture attention in different decoder layers.}
\label{layer}
\end{table}

\subsection{Entropy of Attention Distribution}
\begin{table}[]
\centering
\begin{tabular}{L{0.6cm}|L{0.7cm}|C{0.7cm}|C{1cm}C{1cm}C{1cm}} \toprule[1pt]
\multicolumn{2}{c|}{\multirow{2}{*}{}}           & \multicolumn{4}{c}{\textbf{Entropy}}                                                                                                                                                  \\ \cline{3-6}
\multicolumn{2}{c|}{}              & All  & \begin{tabular}[c]{@{}c@{}}Short\\ {(}0,20{]}\end{tabular} & \begin{tabular}[c]{@{}c@{}}Mid\\ {[}21,40{]}\end{tabular} & \begin{tabular}[c]{@{}c@{}}Long\\ {[}41,$\infty${]}\end{tabular} \\ \midrule
\multicolumn{2}{c|}{\textbf{Baseline}}        & 2.67 & 2.40                                                   & 2.80                                                 & 2.94                                                  \\ \midrule
\multirow{3}{*}{\textbf{Ours}} & GMA         & 0.57 & 0.59                                                   & 0.55                                                 & 0.50                                                  \\
                      & DP & 2.68 & 2.42                                                   & 2.80                                                 & 2.90                                                  \\ 
                      & Total       & 1.86 & 1.78                                                   & 1.90                                                 & 1.90        \\\bottomrule[1pt]                                        
\end{tabular}
\caption{Entropy of attention distribution on varying source sentence length. `GMA': Gaussian mixture attention. `DP': Dot-product attention. `Total': Fusion of two types of attention.}
\label{entropy}
\end{table}

We use Gaussian mixture attention to model concentrated attention to make up for the dispersion of dot-product attention, especially on the long source. Entropy is often used to measure the dispersion of distribution, where the higher entropy means that the distribution is more dispersed \cite{2020arXiv201211747H}. We report the entropy of the attention distribution in our method on varying source sentence length in Table \ref{entropy}. The entropy of dot-product attention increase with the length of the sentence, showing that dot-product attention is easy to become dispersed as the source length increases. However, the entropy of concentrated attention modeled by Gaussian mixture attention always remains at a low level, since it’s unaffected by the source length. Overall, with the proposed Gaussian mixture attention, our method has lower entropy than baseline, indicating that our method focuses more on some important words, which proves to be beneficial to translation \cite{2020arXiv201211747H,8550728}.

\subsection{Alignment Quality}

\begin{table}[]
\centering
\begin{tabular}{L{0.19\textwidth}|C{0.05\textwidth}C{0.05\textwidth}C{0.05\textwidth}} \toprule[1pt]
            & \textbf{AER}     & \textbf{P.} & \textbf{R.}  \\ \midrule
\textbf{{Transformer}-\textit{Base}} & 53.79 & 47.75   & 47.09 \\
\textbf{Our method}  & \textbf{48.21} & \textbf{53.94}   & \textbf{52.83} \\ \midrule
\textbf{{Transformer}-\textit{Big}} & 46.50 & 54.46   & 56.15 \\
\textbf{Our method (\textit{Big})}  & \textbf{43.03} & \textbf{58.86}   & \textbf{60.62} \\\bottomrule[1pt]
\end{tabular}
\caption{Alignment quality of our method and baseline.}
\label{align}
\end{table}

The Gaussian mixture attention we proposed explicitly models the concentrated attention, so it is potential to help cross-attention achieve more accurate alignment between the target and the source. To explore this conjecture, we evaluate the alignment accuracy of our method on RWTH En$\rightarrow$De alignment dataset \footnote{\url{https://www-i6.informatik.rwth-aachen.de/goldAlignment/}} \cite{liu-etal-2016-neural,ghader-monz-2017-attention,tang-etal-2019-understanding}. 

Following \citet{luong-etal-2015-effective} and \citet{kuang-etal-2018-attention}, we force the models to produce the reference target words during inference to get the attention between source and target. We average the attention weights across all heads from the penultimate layer \cite{li-etal-2019-word,ding-etal-2020-self}, where the source token with the highest attention weight is viewed as the alignment of the current target token. The alignment error rate (AER) \cite{och-ney-2003-systematic}, precision (P.), and recall (R.) of our method are reported in Table \ref{align}. 

Our method achieves better alignment accuracy than baseline, improving 5.58 on Transformer-\textit{Base} and 3.47 on Transformer-\textit{Big}, which shows that modeling concentrated attention indeed improves the alignment quality of cross-attention.

\subsection{N-gram Accuracy}

\begin{figure}[t]
\centering
\includegraphics[width=3in]{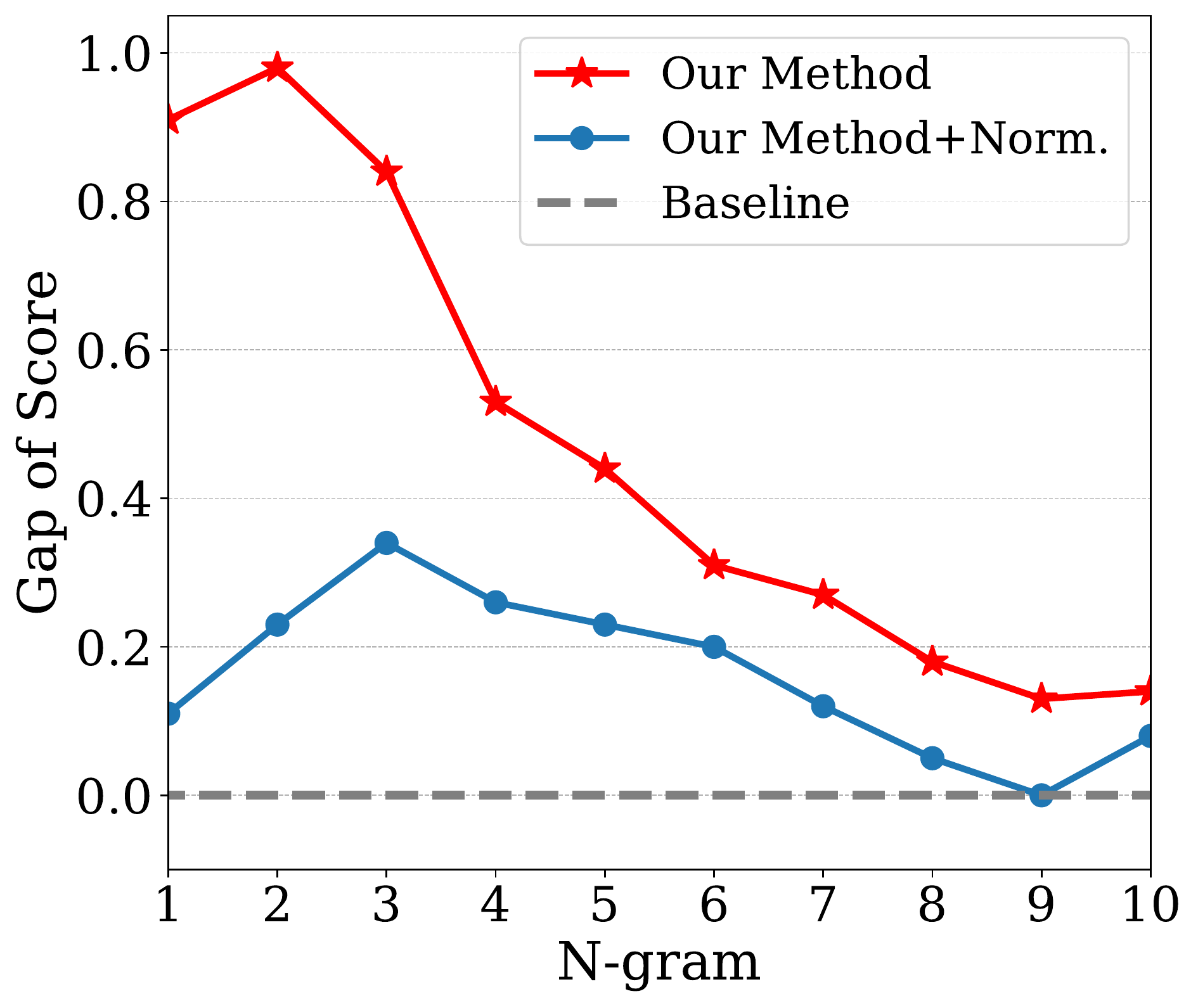}
\caption{Gap of N-gram score between our method and baseline with respect to various N, to highlight the difference in improvement of different n-grams.}
\label{ngram}
\end{figure}

Our method uses Gaussian mixture attention to model concentrated attention, which intuitively should be able to enhance the ability to capture neighboring structures, thereby obtaining more ﬂuent translation. To evaluate the quality of phrase translation, we calculate the improvement of our method on various N-grams in Figure \ref{ngram}. We set Transformer-\textit{Base} as the baseline, and report the gap of score between our method and baseline on each N-gram score.
Our method is superior to the baseline in all N-grams, especially on 2-gram and 3-gram, which shows that our method effectively captures the nearby phrase structure. In the concentrated attention modeled by Gaussian mixture attention, the attention to the surrounding words increases along with the central word, resulting in better phrase translation.

\subsection{Analysis on Sentence Length}

\begin{figure}[t]
\centering
\includegraphics[width=3in]{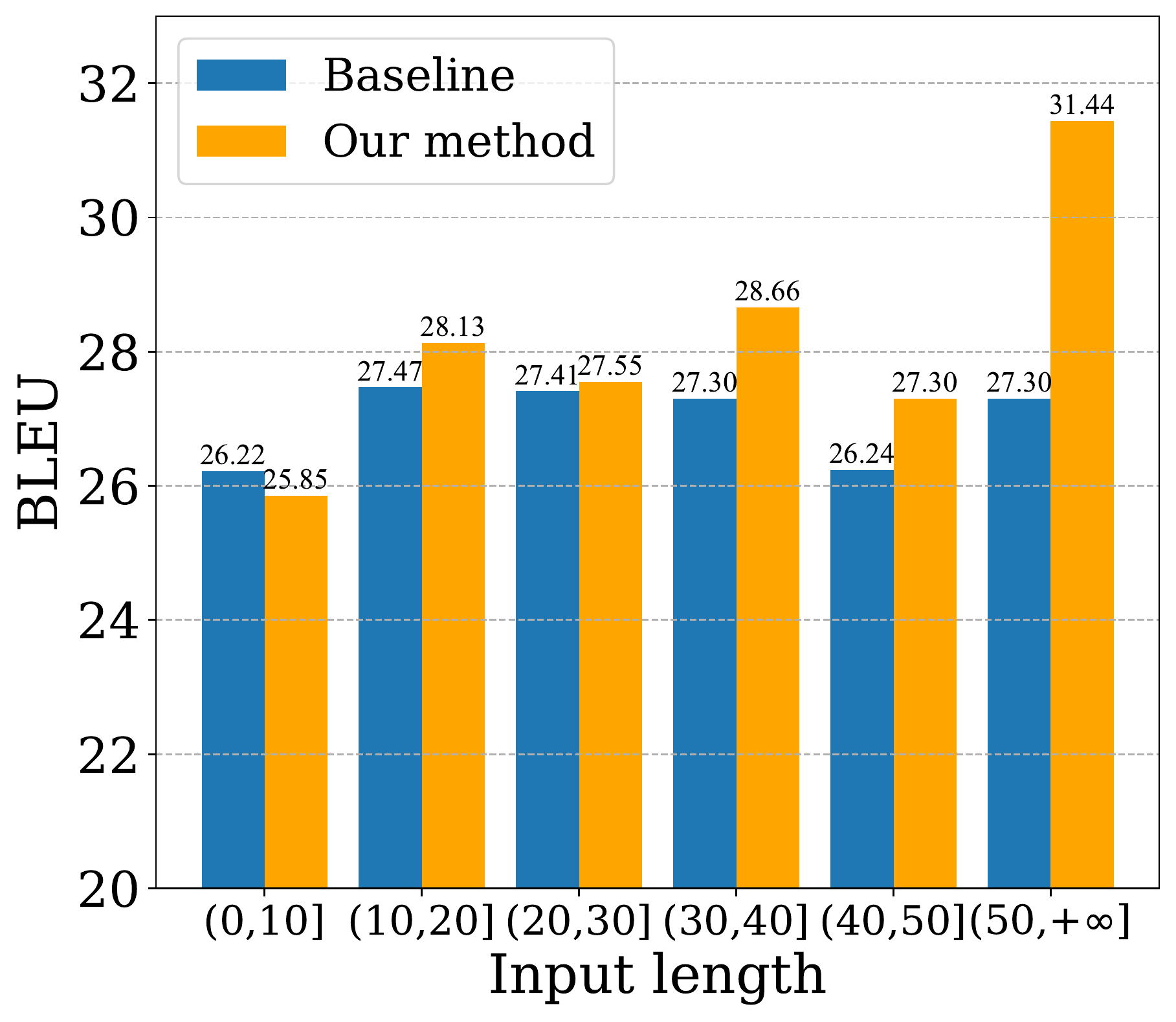}
\caption{BLEU scores of sentence with various length.}
\label{length}
\end{figure}

To analyze the improvement of our method on sentences with different lengths, we group the sentences into 6 sets according to the source length \cite{bahdanau2014neural,tu-etal-2016-modeling}, and report the BLEU scores on each set in Figure \ref{length}. 

Compared with Baseline, our method has a more significant improvement in long sentences, with +1.36 BLEU on $\left (30,40  \right ]$, +1.06 BLEU on $\left (40,50  \right ]$, and +4.14 BLEU on $\left (50,+\infty   \right ]$. Our method significantly improves the long sentence translation by modeling concentrated cross-attention. When the source sentence is very long, dot-product attention fairly pays attention to every source word and normalizes it, causing cross-attention to become dispersed, which proved to be unfavorable for translation in previous work \cite{8550728,2012.11747,tang-etal-2019-understanding}. In contrast, regardless of the length of the source, Gaussian mixture attention concentrates surround some central words, effectively avoiding the attention dispersion. Therefore, our method effectively improves the translation quality of long sentences by modeling the concentrated attention. Besides, our method drops slightly when the sentence length is small. Since we set $K=4$ for the source sentences of different length, when the sentence is very short, it is difficult to accurately find 4 corresponding central words (mean) for each target word.

\section{Conclusion}
Inspired by linguistics, we decompose the cross-attention into distributed attention and concentrated attention. To model the concentrated attention, we apply GMM to construct the Gaussian mixture attention, which effectively resolves the weakness of dot-product attention. Experiments show that the proposed method outperforms the strong baseline on three datasets. Further analyses show the specific advantages of the proposed method in attention distribution, alignment quality, N-gram accuracy, and long sentence translation.

\section*{Acknowledgements}
We thank all the anonymous reviewers for their insightful and valuable comments. This work was supported by National Key R\&D Program of China (NO. 2017YFE0192900).

\bibliography{anthology,custom}
\bibliographystyle{acl_natbib}

\end{document}